\documentclass{llncs}
\usepackage{amsmath,amssymb}
\usepackage{graphicx}
\usepackage{booktabs}
\usepackage{algorithm}
\usepackage{algpseudocode}
\def\twoline#1#2{$\mbox{#1}\atop \mbox{#2}$}
\begin{document}

\title{ Adapting ELM to Time Series Classification: A Novel Diversified  Top-k Shapelets Extraction Method}

\author{Qiuyan Yan\inst{1,2} \and Qifa Sun\inst{1} \and Xinming Yan\inst{1}}

\institute{School of Computer Science and Technology, China University of Mining Technology, Xuzhou, China, 221116
\and School of Safety Engineering, China University of Mining Technology,\\ Xuzhou, China, 221116, \email{yanqy@cumt.edu.cn}}

\maketitle

\begin{abstract} ELM (Extreme Learning Machine) is a single hidden layer  feed-forward network, where the weights between input and hidden layer are initialized randomly. ELM is efficient due to its utilization of the analytical approach to compute weights between hidden and output layer. However, ELM still fails to output the semantic classification outcome. To address such limitation, in this paper, we propose a diversified top-k shapelets transform framework, where the shapelets are the subsequences i.e., the best representative and interpretative features of each class. As we identified, the most challenge problems are how to extract the best k shapelets in original candidate sets and how to automatically determine the k value. Specifically, we first define the similar shapelets and diversified top-k shapelets to construct diversity shapelets graph. Then, a novel diversity graph based top-k shapelets extraction algorithm named as \textbf{DivTopkshapelets}\ is proposed to search top-k  diversified shapelets. Finally, we propose a shapelets transformed ELM algorithm named as \textbf{DivShapELM} to automatically determine the  k value, which is further utilized for time series classification. The experimental results over public data sets demonstrate that the proposed approach significantly outperforms traditional ELM algorithm in terms of effectiveness and efficiency.

\end{abstract}

\section{Introduction}\label{SEC: Introduction}

Extreme Learning Machine (ELM for short) was originally developed based on single hidden-layer feed forward neural networks (SLFNs) \cite{Ref1}. Compared with the conventional learning machines, it is of extremely fast learning capacity and good generalization capability. Thus, ELM, with its variants \cite{Ref2,Ref3}, has been widely applied in many fields \cite{Ref4,Ref5}. The results indicate that ELM produces comparable or better classification accuracies with reduced training time and implementation complexity compared to artificial neural networks methods and support vector machine methods.

Unfortunately, as a black-box method, ELM fails to measure up to the task of time series data classification by itself. A possible solution to this issue is to improve the interpretability of ELM by feature selection. If a set of selected features improve the classification accuracy much more than original feature sets, it is reasonable to interpret the results by them. Nevertheless, selection the most representative and interpretative feature can improve the interpretability of ELM and make ELM more adapt to time series classification. In the context of feature selection in time series data analysis, most of the current methods adopt such a framework that ranks subsequences according to their individual discriminative power to the target class and then selects top-k ranked subsequences\cite{Ref6}. These methods have some common drawbacks: (1) the selected features are not the most representative and interpretative, (2) many redundant features are selected, and (3) the number of selected feature is arbitrarily specified by a parameter k .

In this paper, a novel method is proposed to improve ELM representative and interpretative ability by extraction diversified top-k shapelets features.Shapelets was introduced as a primitive for time series data mining \cite{Ref7} and was utilized in classifying time series data \cite{Ref8,Ref9}. The original shapelet based classifier embeds the shapelet discovery algorithm in a decision tree, and uses information gain to assess the quality of candidates. Shapelets transformed classification methods \cite{Ref10,Ref11} were proposed to separate the processing of shapelets selection and classification. The k shapelets are selected in an offline manner, and which can not only improve the affectivity and efficiency of classification, but also introduce a common feature attraction method which can be used in all typical time series classification algorithms. Nevertheless, shapelets based classification methods have been widely discussed and used in many real applications \cite{Ref12,Ref13,Ref14}.

The most challenge is that there are large quantities of redundant shapelets in candidates which decreasing the accuracy of classification and the parameter k is hard to determine. Some works \cite{Ref11,Ref15} detect this problem and use clustering or pruning methods to remove the redundant,but still exist redundant shapelets, also, the k value is determined from experiments. This paper make the following contributions: First, in order to get rid of the similar and redundant shapelets in candidate set, two conceptions including similar shapelets and diversified top-k shapelets are presented. Based on these conceptions, a method of construction diversify shapelets graph is proposed. Second, a diversified top-k shapelets query method is presented to find top-k  representative shapeletes of each class. Third, we propose an diversified top-k shapelets transformed ELM algorithm which can automatically determine the parameter k and transform data using the determined k shapelets. The experimental results show that the proposed approach significantly improves the interpretability and performance of ELM.

\section{Preliminary}\label{SEC: Preliminary}

Extreme Learning Machine (ELM) is a generalized single hidden-layer feedforward network. In ELM, the hidden layer node parameters are mathematically calculated instead of being iteratively tuned; thus, it provides good generalization performance at thousands of times faster speed than traditional popular learning algorithms for feedforward neural networks.

Suppose there are $ N $ arbitrary distinct training instances $({x_i},{t_i})$, where ${x_i} = {[{x_{i1}},{x_{i2}},...,{x_{in}}]^T} \in {{\rm{R}}^n}$, and ${t_i} = {[{t_{i1}},{t_{i2}},...,{t_{in}}]^T} \in {{\rm{R}}^m}$,standard SLFNs with $\tilde N$ hidden nodes and activation function $g(x)$ are mathematically modeled as \begin{equation} \sum\limits_{i = 1}^{\tilde N} {{\beta _i}{g_i}({x_j})}  = \sum\limits_{i = 1}^{\tilde N} {{\beta _i}{g_i}({{\rm{w}}_i} \cdot {x_j} + {b_i})}  = {{\rm{o}}_j},j = 1,...,N,\label{eq:1}\end{equation} where ${{\rm{w}}_i} = {[{w_{i1}},{w_{i2}},...,{w_{in}}]^T}$ is the weight vector connecting the $i$th hidden node and the input nodes, ${\beta _i} = {[{\beta _{i1}},{\beta _{i2}},...,{\beta _{im}}]^T}$ is the weight vector connecting the $i$th hidden nodes and the output nodes, and ${b_i}$ is the threshold of the $i$th hidden node.If a SLFN with ${\tilde N}$ hidden nodes with activation function $g(x)$ can approximate these $ N $ samples with zero error, it then implies that there exist $ {\beta _i} $, $ {w_i} $ and $ {b_i} $, such that:
\begin{equation} \sum\limits_{i = 1}^{\tilde N} {{\beta _i}} G({w_i} \cdot {x_j} + {b_i}) = {t_j},j = 1, \ldots ,N, \label{eq:2}\end{equation}
The above $N$ equations can be written compactly as
\begin{equation} \operatorname{H} \beta  = \operatorname{T},  \label{eq:3}\end{equation}
where
$$ \begin{gathered}
\begin{array}{l}
{\mathop{\rm H}\nolimits} ({w_1}, \ldots ,{w_{\tilde N}},{b_1}, \ldots ,{b_{\tilde N}},{x_1}, \ldots ,{x_N})\\
 = {\left( {\begin{array}{*{20}{c}}
{g({w_1} \cdot {x_1} + {b_1})}& \ldots &{G({w_{\tilde N}} \cdot {x_1} + {b_{\tilde N}})}\\
 \vdots & \ddots & \vdots \\
{g({w_1} \cdot {x_N} + {b_1})}& \cdots &{G({w_{\tilde N}} \cdot {x_N} + {b_{\tilde N}})}
\end{array}} \right)_{N \times \tilde N}}
\end{array}
\end{gathered} $$
$$ \mbox{ $\beta  = \left[ \begin{gathered}
\beta _1^T \hfill \\
\ldots \hfill \\
\ldots \hfill \\
\beta _{\tilde N}^T \hfill \\
\end{gathered}  \right]_{\tilde N \times m} $ and $ \operatorname{T}  = {\left[ \begin{gathered}
t_1^T \hfill \\
\ldots \hfill \\
\ldots \hfill \\
t_N^T \hfill \\
\end{gathered}  \right]_{N \times m}} $ } $$

H is named as hidden layer output matrix of the network, where with respect to inputs $ {x_1},{x_2},\ldots,{x_N} $ and its $j$th row represents the output vector of the hidden layer with respect to input $ {x_j} $.

ELM differs from other training algorithms in that the hidden node parameters $ {w_i} $ and $ {b_i} $ are not tuned during training, but are instead assigned with random values according to any continuous samplings distribution. Eq. \eqref{eq:3} then becomes a linear system and the output weight $ \beta  $ are estimates as Eq. \eqref{eq:4}.
\begin{equation} \overset{\lower0.5em\hbox{ $\smash{\scriptscriptstyle\frown} $ }}{\beta }  = {\operatorname{H} ^\dag }T, \label{eq:4}\end{equation}
where $ {\operatorname{H} ^\dag } $ is the Moore-Penrose generalized inverse of the hidden layer output matrix H.

\section{Diversified Top-k Shapelets Transformed ELM}\label{SEC: Diversified Top-k Shapelets Transformed ELM}

In this section, we discuss three parts of our work (1) construction the diversity graph of shapelets candidates, (2) querying diversified top-k shapelets, and (3) transforming the data based on diversified top-k shapelets and applying in ELM. The following contents will discuss above three contribution separately.

Before removing the redundant shapelets, we firstly need to get the shapelets candidates set. The original shapelets extraction algoritm is time consuming and complexity is $ O({n^2}{m^4}) $, $n$ is the number of time series in the data set, $m$ is the length of each time series. In order to improve the  efficiency of shapelets based classification method, we follow the method proposed in \cite{Ref9}, which transformed the data sets through SAX method and decreased the time complexity to $ O(n{m^2}) $.

\subsection{Construction the diversity graph of shapelets candidates}\label{SSEC: Construction the diversity graph of shapelets candidates}

Considerable works have focused on the diversified top-k query, but they almost applying on a typical circumstance. In our work, we use the diversity graph\cite{Ref16} to find a general method to extract diversified top-k shapelets.

Given $I$ is a shapelets candidate sets, $ I = \{ {s_1},\ldots {\mkern 1mu} {s_n}\} $, and $n$ is the number of $I$. The question is how to measure the similarity of two shapelets and how to define the diversified top-k shapelets. So we first give the two definitions.

\textbf{Definition 1: Similar shapelets.}
Given two shapelets $ {s_i} $ and $ {s_j} $ which represent the same class, $ 1 \le i,j \le n,i \ne j $ and $ n $ is the number of shapelets candidates. The optimal split point of $ {s_i} $ and $ {s_j} $ are $  < {s_i},{d_i} >  $ and $  < {s_j},{d_j} >  $, the split threshold are $ {d_i} $ and $ {d_j} $. We say $ {s_i} $ and $ {s_j} $ are similar shapelets when they satisfy $ dis({s_i},{s_j}) \le \min ({d_i},{d_j}) $. We denote the similar shapelets as $ {s_i} \approx {s_j} $.

\textbf{Definition 2: Diversified top-k Shapelets.} Given a shapelets candidates set $ I = \{ {s_1},\ldots {\mkern 1mu} {s_n}\} $, and an integer k where $ 1 \le k \le |I| $. The diversified top-k shapelets query results of $ I $, denoted as DivTopk($ I $), is a list of results that satisfy the following three conditions.

1) DivTopk($ I $) $ \subseteq $ I, $ | $DivTopk($ I $)$ | \le k $

2) For any two results $ {s_i} \in I $ and $ {s_j} \in I $ and $ {s_i} \ne {s_j} $, if $ {s_i} \approx {s_j} $, then $ \{ {s_i},{s_j}\}  \not\subset $DivTopk($ I $).

3) $ \sum\limits_{si \in I} {\mathrm{score}({s_i})}  $ is maximized.

\begin{algorithm}
  \caption{conShapeletGraph(allShapelets)}
  \textbf{input:} shapelets candidates allShapelets\\
  \textbf{output:} diverisity shapelets Graph
  \begin{algorithmic}[1]
    \State Graph = $ \phi  $
    \State sort(allShapelets)
    \For{i=1 to $ | $allShapelets$ | $}
      \State Graph.add(allShapelets[i])
    \EndFor
    \For{j=1 to $ | $allShapelets$ | $}
      \For{k=1 to $ | $allShapelets$ | $}
        \If{(allShapelets[j] $\approx $ allShapelets[k])}
          \State Graph[j].add(Graph[k])
          \State Graph[k].add(Graph[j])
        \EndIf
      \EndFor
    \EndFor
    \State \textbf{return} Graph
  \end{algorithmic}
\end{algorithm}

We give a diversity shapelet graph example of ChlorineConcentration dataset as in Fig. 1.There are ten shapelets candidates as shown in fig1-a and the diversity graph of these ten candidates from algorithm 1 are shown in fig1-b.The black, red and green subsequence are the top-3 shapelets and can get the best classification accuracy.Next section we will explain how to get the diversified top-k shapelets on the diversify graph.

\begin{figure}[!htb]
    \centering
    \begin{minipage}[b]{.48\linewidth}
      \centering
      \includegraphics[scale=1]{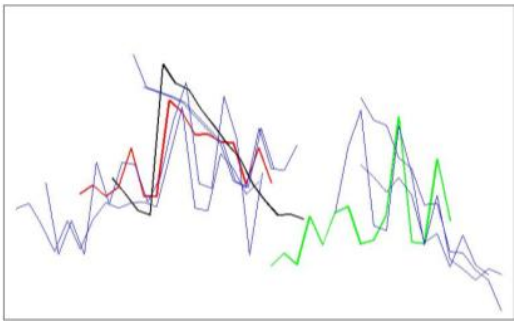}\\
      (a) shapelets candidates
    \end{minipage}
    \hfill
    \begin{minipage}[b]{.48\linewidth}
      \centering
      \includegraphics[scale=1]{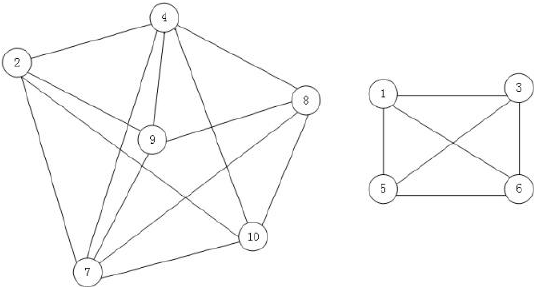}\\
      (b) diversity graph
    \end{minipage}
    \caption{\label{fig:1}example of diversity shapelets graph}
\end{figure}

\subsection{Diversified Top-k Shapelets Extraction}\label{SSEC: Diversified Top-k Shapelets Extraction}

Traditional top-k query only returns the objects with largest k score, however, diversified top-k query concerns not only the score value but also the similarity of each object and remove all the redundant objects from results. According to \cite{Ref16}, find top-k results falling into two categories: incremental manner and bounding manner. We noticed that the bounding manner first satisfied the k value, but in each step it may not add the largest score vertex. In our problem, we must maintain the largest information gain shapelets in order to have the best classification accuracy. So we calculate the diversified top-k shapelets via an incremental manner. The detailed search procedure is shown as in algorithm2.

\begin{algorithm}
  \caption{DivTopkShapelets (Graph, k)}
  \textbf{input:} diverisity Graph, k value\\
  \textbf{output:} k number of shapelets
  \begin{algorithmic}[1]
    \State kShapelets = {\O}, n = $ | $V(Graph)$ | $
    \State kShapelets.add(v1)
    \State while($ | $ kShapelets $ | $$ < $ k)
    \For{i=2 to n}
      \If{(Graph[i] $\cap $ kShapelets = {\O})}
        \State kShapelets.add(vi)
      \EndIf
    \EndFor
    \State \textbf{return} kShapelets
  \end{algorithmic}
\end{algorithm}

\subsection{Diversified Top-k Shapelets Transform for ELM}\label{SSEC: Diversified Top-k shapelets Transform for ELM}

After getting diversified top-k shapelets, we can use these shapelets to transform data before ELM classification. For each instance of data ${T_i}$ , the subsequence distance is computed between ${T_i}$ and ${S_j}$ , ${S_j}$ is a shapelet in Top-k shapelets.The resulting $k$ distances are used to form a new instance of transformed data, where each attribute corresponds to the distance from each shapelet to the original time series.

In order to get the best classification accuracy and also to get rid of the independence on the parameter k, we set k in an interval of [1, $ \kappa  $ ] where $ \kappa  $ is an empirical optimal value, according to our experiments(see section  4.1), which is set to 9,then we use the ELM to learning training data and evaluate each diversified top-k shapelets candidate. The k value with the largest prediction accuracy is selected.

When using data split into training data and testing data, the shapelets extraction and k determination is carried out only on the training data to avoid bias. The optimal diversified top-k shapelets are then used to transform each instance of the testing data.The details are as in following algorithm 3.

\begin{algorithm}
  \caption{DivShapELM(Graph, $ \kappa $)}
  \textbf{input:} $ \kappa  $ value\\
  \textbf{output:} ELM classification results
  \begin{algorithmic}[1]
    \For {i=1 to $ \kappa $}
      \State kShapelets = DivTopkShapelet(Graph,k)
      \State output = {\O}
      \For{ts= 1 to $\left| {{\rm{Dt}}} \right|$}
          \State transformed = {\O}
          \For { s = 1 to $\left| {{\rm{kshapelets}}} \right|$}
              \State dist = subsequenceDist(ts,s)
              \State transformed.add(dist)
          \EndFor
          \State output.add(transformed)
      \EndFor
      \State using ELM evaluate output
      \State kShapelets = the output with highest accuracy on ELM
    \EndFor
    \State using kShapelets transform testing data
    \State \textbf{return} ELM classification results
  \end{algorithmic}
\end{algorithm}

\section{Experiments}\label{SEC: Experiments}

To evaluate our proposed methods, we selected 15 data sets from the UCR time series repository (listed in Table 1).We use a simple train/test split and all reported results are testing accuracy.All shapelets candidates seletction, top-k diversified shapelets extraction and classifier construction is done on the training set. All experiments are implemented in Java within the Weka framework.

\subsection{Determination of shapelets length and  $ \kappa $}\label{SSEC: Determination of shapelets length and k}

There are two parameters \textit{min} and \textit{max} in the procedure of shapelets candidates generation. The two parameters determine the length of shapelets candidates which can influence finding the best representative shapelets. Followed \cite{Ref11}, we set min-length and max-length of subsequences to generate shapeless are m/11 and m/2 separately, m is the length of each time series.

First, in order to explain how the k value influences accuracy of classification, we test the average accuracy of six classifier on fifteen data sets with the varying k value. As shown in Fig. 2, with the increasing of k value, average classification accuracy first increases and then becomes stable when k is 9. Accordingly, we set the $ \kappa  $ value as 9 and use this value in the following experiments.

\begin{figure}[!htb]
    \centering
    \includegraphics[scale=1]{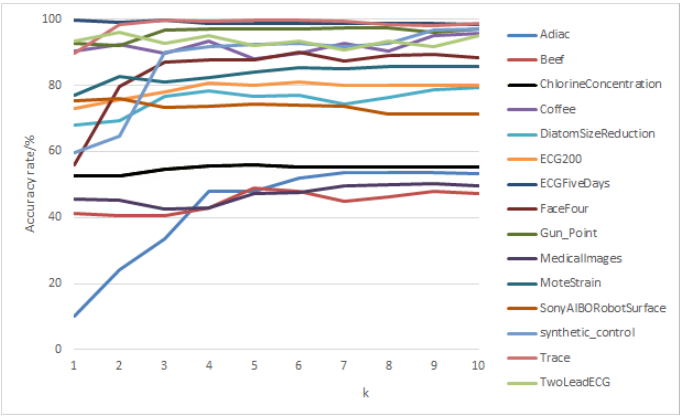}\\
    \caption{\label{fig:2}Accuracy varying with k}
\end{figure}

\subsection{Representation of Optimal Shapelets Sets}\label{SSEC: Representation of optimal shapelets}

In this section, we want to get a visual overlook at what was the optimal shapelets indeed. Because reference\cite{Ref15} has verified that ShapeletSelection can remove more redundant shapelets than other similar methods, we only compared the optimal shapelets sets between \textbf{DivTopkShapelets} and ShapeletSelection when the two algorithms all have the best classification,as shown in Fig. 3. The optimal shapelets sets were acquired from ShapeletSelection when k=8 (Fig.3-a), and from DivTopkShapelet when k=2(Fig.3-b).

\begin{figure}[!htb]
    \centering
    \begin{minipage}[b]{.48\linewidth}
      \centering
      \includegraphics[scale=1]{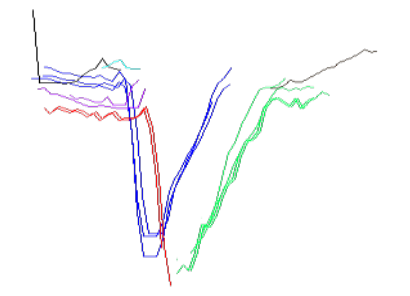}\\
      (a) optimal 8 shapelets of ShapeletSelection with the best accuracy
    \end{minipage}
    \hfill
    \begin{minipage}[b]{.48\linewidth}
      \centering
      \includegraphics[scale=1]{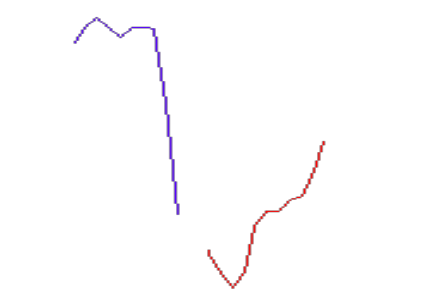}\\
      (b) optimal 2 shapelets of DivTopkShapelet with the best accuracy
    \end{minipage}
    \caption{\label{fig:3}Optimal shapelets sets}
\end{figure}
\subsection{Accuracy Comparison}\label{SSEC: Accuracy Comparison}
In this section, we select six traditional time series classification algorithms including C4.5, 1NN, Naive Bays(NaB), BayesianNetwork(BaN), RandomForest(RaF) and RotationForest(RoF) to compare the accuracy with our proposed methods.

\subsubsection{Accuracy comparison with Traditional Classification}\label{SSSEC: Accuracy comparison with Traditional Classification}

Firstly, we directly use selected classification algorithms to classify the datasets .Secondly,we use \textbf{DivTopkShapelets}(set k=9)to extract optimal shapeletes sets and transform data, then classify transformed data sets with six selected classification algorithms.Results are presented in table 1, the column captions with classifier name plus S(see C4.5(S)) means \textbf{DivTopkShapelets} transformed classification results. From the table we can see that compared to traditional classification algorithms, \textbf{DivTopkShapelets} transformed classification methods can improve accuracy of 9 out of 15 datasets. For all six classification algorithms, average accuracy are improved. Especially for NaiveBays, \textbf{DivTopkShapelets} improves 13 data sets accuracy.

\begin{table}[!htb]
    \centering
    \caption{\label{tab:1}Accuracy comparison with traditional classification algorithm(The method/s
with the highest accuracy in each database are shown in bold)}
    \renewcommand\tabcolsep{6pt}
    (a)\\
    \begin{tabular}{lrrrrrr}
      \toprule
      \multicolumn{1}{c}{Data} & \multicolumn{1}{c}{C4.5} & \multicolumn{1}{c}{C4.5(S)} & \multicolumn{1}{c}{1NN} & \multicolumn{1}{c}{1NN(S)} & \multicolumn{1}{c}{NaB} & \multicolumn{1}{c}{NaB(S)} \\
      \midrule
      Adiac & 53.19 & 49.36 & 59.34 & 56.27 & 56.52 & 57.54 \\
      Beef  & 56.67 & 40    & 60    & 53.33 & 50    & 60 \\
      Chlorine & 64.3  & 56.82 & 68.52 & 58.59 & 34.61 & 45.52 \\
      Coffee & 57.14 & 92.86 & 75    & \textbf{100}   & 67.86 & 92.86 \\
      Diatom & 71.24 & 67.65 & 93.46 & \textbf{94.44} & 87.91 & 78.76 \\
      ECG200 & 72    & 79    & \textbf{89}    & 78    & 77    & 80 \\
      ECGFiveDays & 72.12 & 98.61 & 80.6  & \textbf{99.77} & 79.67 & 96.4 \\
      FaceFour & 71.59 & 78.41 & 87.5  & \textbf{97.73} & 84.09 & 82.95 \\
      Gun\_Point & 77.33 & 92    & 92    & 96.67 & 78.67 & 95.33\\
      MedicalImages & 62.5  & 49.08 & 67.89 & 46.45 & 44.87 & 51.84 \\
      MoteStrain & 78.67 & 78.67 & 85.78 & 89.62 & 84.19 & 85.7 \\
      SonyAIBORobot & 65.56 & 92.51 & 67.72 & 95.51 & 92.85 & 95.51 \\
      synthetic\_control & 81    & 95    & 88    & \textbf{98}    & 96    & 96.33 \\
      Trace & 74    & \textbf{100}   & 82    & 98    & 80    & 95 \\
      TwoLeadECG & 71.82 & 91.22 & 72.52 & 98.95 & 69.8  & \textbf{99.65} \\
      average & 68.61 & 77.41 & 77.95 & \textbf{84.09} & 72.28 & 80.89 \\
      improve datasets &     & 10      &     & 10      &     & 13 \\
      \bottomrule
    \end{tabular}\\
    (b)\\
    \begin{tabular}{lrrrrrr}
      \toprule
      \multicolumn{1}{c}{Data} & \multicolumn{1}{c}{BaN} & \multicolumn{1}{c}{BaN(S)} & \multicolumn{1}{c}{RaF} & \multicolumn{1}{c}{RaF(S)} & \multicolumn{1}{c}{RoF} & \multicolumn{1}{c}{RoF(S)} \\
      \midrule
      Adiac & 50.9  & 38.87 & 62.15 & 58.82 & \textbf{62.29} & 60.1 \\
      Beef  & 60    & 33.33 & 53.33 & 46.67 & \textbf{80} & 53.33 \\
      Chlorine & 59.9  & 56.09 & 70.76 & 57.68 & \textbf{81.93} & 57.5 \\
      Coffee & 64.29 & 96.4 & 64.29 & 92.86 & 94.12 & 96.43\\
      Diatom & 94.12 & 77.45 & 88.89 & 75.16 & 86.93 & 78.1 \\
      ECG200 & 75    & 81    & 82    & 82    & 83    & 80 \\
      ECGFiveDays & 78.05 & 98.26 & 68.99 & 99.19 & 90.71 & 99.65 \\
      FaceFour & 89.37 & 88.64 & 87.5  & 92.05 & 75    & 96.59 \\
      Gun\_Point & 85.33 & \textbf{99.33} & 96    & 96    & 86    & 98 \\
      MedicalImages & 40.13 & 47.37 & 71.71 & 53.03 & \textbf{72.37} & 53.16 \\
      MoteStrain & 85.62 & 85.78 & 85.98 & 84.5  & 84.82 & \textbf{89.94} \\
      SonyAIBORobot & 74.04 & \textbf{95.84} & 71.38 & 95.67 & 72.88 & 95.51 \\
      synthetic\_control & 92.67 & 96.67 & 93.67 & 96.33 & 92.67 & 97.67\\
      Trace & 82    & \textbf{100}   & 80    & \textbf{100}   & 91    & 96 \\
      TwoLeadECG & 73.22 & 97.98 & 71.73 & 93.42 & 91.66 & 94.73 \\
      average & 73.64 & 79.54 & 76.56 & 81.56 & 83.03 & 83.11 \\
      improve datasets &    &10        &      & 9      &      & 9 \\
      \bottomrule
    \end{tabular}
\end{table}

\subsubsection{Accuracy comparison with ShapeletSelection Algorithm}\label{SSSEC: Accuracy comparison with ShapeletSelection method}

We compared the relative accuracy of \textbf{DivTopkShapelets} with the most similar method ShapeletSelection as shown in table2. From results we can draw the conclusion that compared with ShapeletSelection, \textbf{DivTopkShapelets} transformed classification method can improve accuracy of 9 out of 15 datasets. On Adiac dataset, \textbf{DivTopkShapelets} has the best performance, the average accuracy improved 20.08\%. For classifiers, \textbf{DivTopkShapelets} increase  1NN classifier most with the accuracy improved 6.06\%.

\subsubsection{Accuracy Comparison with ELM}\label{SSSEC: Accuracy comparison between DivTopkELM and ELM}

In this section,we compare the classification accuracy between \textbf{DivShapELM} and traditional ELM and average accuracy of six traditional classification algorithms(column named as $ {\rm{avg\_T}} $). As shown in table 3,\textbf{DivShapELM} can obvious improve traditional ELM accuracy. In 13 out of 15 datasets, the accuracy of ELM is improved and is close to or even better than average accuracy of  traditional classification algorithms. Especially, in Trace dataset, \textbf{DivShapELM} has the accuracy of 99.20, better then ELM 39.40\%.

\begin{table}[!htb]
    \centering
    \caption{\label{tab:2}Relative accuracy between \textbf{DivTopkShapelets} and ShapeletSelection(The highest average relative accuracy in each database and each classifier are shown in bold.The negative value means accuracy are not improved. ) )}
    \begin{tabular}{lrrrrrrr}
      \toprule
      \multicolumn{1}{c}{Data} & \multicolumn{1}{c}{C4.5(S)} & \multicolumn{1}{c}{1NN(S)} & \multicolumn{1}{c}{NaB(S)} & \multicolumn{1}{c}{BaN(S)} & \multicolumn{1}{c}{RaF(S)} & \multicolumn{1}{c}{RoF(S)} & \multicolumn{1}{c}{Average} \\
      \midrule
      Adiac & 16.62 & 21.23 & 17.14 & 16.37 & 26.09 & 23.02 & \textbf{20.08} \\
      Beef  & -3.33 & 3.33  & 16.67 & 0.00  & 3.33  & 10.00 & 5.00 \\
      Chlorine & 0.10  & 14.24 & -10.70 & -0.65 & 13.54 & 0.78  & 2.89 \\
      Coffee & -3.57 & 7.14  & 0.00  & 0.00  & -7.14 & 7.14  & 0.59 \\
      Diatom & 5.56  & 9.15  & -6.86 & -4.90 & -16.34 & -5.89 & -3.21 \\
      ECG200 & 4.00  & -6.00 & 2.00  & 3.00  & 5.00  & 0.00  & 1.33 \\
      ECGFiveDays & -0.46 & -0.23 & -1.28 & -0.58 & -0.35 & 0.70  & -0.37 \\
      FaceFour & 3.41  & 0.00  & 1.13  & -7.95 & -1.13 & 9.09  & 0.76\\
      Gun\_Point & 0.67  & -2.00 & -1.34 & -0.67 & -3.33 & -0.67 & -1.22 \\
      MedicalImages & 1.84  & 8.55  & 0.53  & -4.08 & 15.00 & 5.92  & 4.63 \\
      MoteStrain & -4.63 & 1.77  & -5.11 & -5.59 & -6.15 & -1.03 & -3.46 \\
      SonyAIBORobot & 16.97 & 32.61 & 5.99  & 8.49  & 12.64 & 11.32 & 14.67 \\
      synthetic\_control & 3.00  & 1.67  & -0.67 & 0.67  & -1.00 & 0.00  & 0.61 \\
      Trace & 6.00  & 0.00  & -4.00 & 2.00  & 0.00  & -2.00 & 0.33 \\
      TwoLeadECG & -3.07 & -0.52 & 0.88  & -0.35 & -6.05 & -2.46 & -1.93 \\
      Average  & 2.87  & \textbf{6.06} & 0.96  & 0.38  & 2.27  & 3.73  & 2.71 \\
      Data sets improved & 10    & 11    & 8     & 7     & 7     & 10    & 10 \\
      \bottomrule
    \end{tabular}
\end{table}

\begin{table}[!htb]
    \centering
    \caption{\label{tab:3}Accuracy comparison between \textbf{DivShapELM} and rival methods(The method/s
with the highest accuracy in each database are shown in bold))}
    \renewcommand\tabcolsep{10pt}
    \begin{tabular}{lrrrrr}
      \toprule
      \multicolumn{1}{c}{Data} & \multicolumn{1}{c}{ELM} & \multicolumn{1}{c}{DivShapELM} & \multicolumn{1}{c}{ $ {\rm{avg\_T}} $}& \\
      \midrule
      Adiac & 30.44 & 43.52  & \textbf{57.40}  \\
      Beef  & \textbf{70.67} & 44.00 & 60.00  \\
      Chlorine & 54.95 & 57.35 & \textbf{63.34}  \\
      Coffee & 61.79 & \textbf{92.15}  & 70.45  \\
      Diatom & 70.43 & 64.31 & \textbf{87.09}  \\
      ECG200 & 78.60 & 79.60   & \textbf{79.67}  \\
      ECGFiveDays & 70.12 & \textbf{97.45}  & 78.36  \\
      FaceFour & 43.18 & \textbf{89.54} & 82.51  \\
      Gun\_Point & 83.60 & \textbf{95.80}   & 85.89  \\
      MedicalImages & 54.42 & 56.72  & \textbf{59.91}  \\
      MoteStrain & 58.33 & 69.08   & \textbf{84.18}  \\
      SonyAIBORobot & 54.71 & 71.68   & \textbf{74.07}  \\
      synthetic\_control & 65.23 & \textbf{97.23} & 90.67  \\
      Trace & 59.80 & \textbf{99.20}   & 81.50  \\
      TwoLeadECG & 67.69 & \textbf{78.68}   & 75.13  \\
      \bottomrule
    \end{tabular}
\end{table}

\subsection{Runtime comparison}\label{SSEC: Time cost comparison}

\textbf{DivShapELM} has three extra pre-procedures: shapelets candidate selection, diversified shapelets selection and data transform. Once the transformed data are prepared, the rest procedure is a usual classification process. Table 4 gives the extra time and classification time of \textbf{DivShapELM}  and ELM. The time cost of diversified shapelets selection are varied with data sets, but which can be conducted in an offline manner. Because \textbf{DivShapELM}  can transform a dataset from $ n \times m $ length to a ${{\bf{R}}^{n \times k}}$ matrix(k $ \ll $ m), the runtime of \textbf{DivShapELM} can be reduced larglely. As shown in table 4, \textbf{DivShapELM}  has the less classification time on 12 out of 15 datasets.

\begin{table}[!htb]
    \centering
    \caption{\label{tab:4}Runtime of \textbf{DivShapELM}  and ELM (The method with minimum runtime in each database  are shown in bold and time in seconds)}
    \renewcommand\tabcolsep{5pt}
    \begin{tabular}{lrrrrr}
      \toprule
      \multicolumn{1}{c}{Data}  & \multicolumn{1}{c}{\twoline{candidate}{selection}} & \multicolumn{1}{c}{\twoline{Diversified}{top-kshapelets}} & \multicolumn{1}{c}{\twoline{Data}{transform}}  & \multicolumn{1}{c}{DivShapELM} & \multicolumn{1}{c}{ELM} \\
      \midrule
      Adiac & 1277  & 18.81  & 0.811 & \textbf{0.04992} & 0.13728 \\
      Beef  & 1026  & 2267.046 & 0.702 & 0.0156 & 0.0156 \\
      Chlorine & 2636  & 28.224 & 7.363 & \textbf{0.01716} & 0.05148 \\
      Coffee & 337   & 2109.798 & 0.436 & 0.04368 & 0.00936 \\
      Diatom & 95    & 4286.979 & 2.294 & \textbf{0.02652} & 0.0312 \\
      ECG200 & 216   & 8.145 & 0.14  & \textbf{0.0156} & 0.02496 \\
      ECGFiveDays & 84    & 7.722 & 0.717 & \textbf{0.02184} & 0.02652 \\
      FaceFour & 634   & 602.46 & 0.671 & \textbf{0.02184} & 0.06864 \\
      Gun\_Point & 151   & 28.503 & 0.218 & \textbf{0.0312} & 0.09048 \\
      MedicalImages & 732   & 1.971 & 0.514 & \textbf{0.00468} & 0.05304 \\
      MoteStrain & 30    & 6.183 & 0.483 & \textbf{0.03276} & 0.04368 \\
      SonyAIBORobot & 28    & 3.51  & 0.249 & 0.01404 & 0.00624 \\
      synthetic\_control & 340   & 3.087 & 0.25  & \textbf{0.04368} & 0.07488 \\
      Trace & 1168  & 2413.485 & 1.201 & \textbf{0.01716} & 0.1638 \\
      TwoLeadECG & 25    & 2.385 & 0.546 & \textbf{0.039} & 0.1017 \\
      \bottomrule
    \end{tabular}
\end{table}

\section{Conclusion and Future Work}\label{SEC: Conclusion}

In this paper,we  proposed a novel method to adapt ELM to time series classification by extraction diversified top-k shapelets. Our work includes three parts: (1) we introduce two conceptions of similar shapelets and diversified top-k shapelets, based on these conceptions, a method of construction diversity shapelets graph is presented, (2) we propose a diversified top-k shapelets extraction method, named as \textbf{DivTopkShaplete}, to find out all of the most representative and interpretative features of each class, and (3) we put forward a shapelets transformed ELM algorithm, named as \textbf{DivShapELM}, which automatically determine k value and get the diversified top-k shapelets to improve performance of ELM. The experiments results show that \textbf{DivShapELM} can improve the efficiency and interpretative of ELM . Also, we experimentally verify that \textbf{DivTopkShaplete} is an excellent feature extraction method which can improve the accuracy of traditional time series classification algorithms.

For future work, we plan to leverage multi-view feature representations \cite{Ref17,Ref18,Ref19,Ref20,Ref21,Ref22,Ref23} to achieve the performance improvement.

\end{document}